%% file: a_main.tex
\documentclass[lettersize,journal]{IEEEtran}
\usepackage{amsmath,amsfonts}
\usepackage{algorithmic}
\usepackage{algorithm}
\usepackage{array}
\usepackage[caption=false,font=normalsize,labelfont=sf,textfont=sf]{subfig}
\usepackage{textcomp}
\usepackage{stfloats}
\usepackage{url}
\usepackage{verbatim}
\usepackage{graphicx}
\usepackage{cite}
\usepackage{abbreviation}
\usepackage[colorlinks=true, linkcolor=blue, citecolor=blue, urlcolor=blue]{hyperref}
\usepackage[table]{xcolor}  
\usepackage{amsmath}
\usepackage{amssymb}
\usepackage{makecell}       
\usepackage{adjustbox}
\usepackage{graphicx}       
\usepackage{multirow}       
\usepackage{booktabs}       
\usepackage[table]{xcolor}
\usepackage{colortbl}
\usepackage{threeparttable}
\usepackage{pifont}

\definecolor{deepred}{RGB}{139,0,0}
\definecolor{deepgreen}{RGB}{0,100,0}

\begin{document}

\title{YOLOv13: Real-Time Object Detection with Hypergraph-Enhanced Adaptive Visual Perception}

\author{Mengqi~Lei, 
        Siqi~Li, 
        Yihong~Wu,  
        Han~Hu, 
        You~Zhou, 
        Xinhu~Zheng, 
        Guiguang~Ding,~\IEEEmembership{Senior Member,~IEEE},
        Shaoyi~Du,~\IEEEmembership{Member,~IEEE}, 
        Zongze~Wu,~\IEEEmembership{Member,~IEEE}, 
        Yue~Gao,~\IEEEmembership{Senior Member,~IEEE}\vspace{-0.2cm}
\IEEEcompsocitemizethanks{
\IEEEcompsocthanksitem Mengqi Lei, Siqi Li, Guiguang Ding, and Yue Gao are with BNRist, THUIBCS, BLBCI, School of Software, Tsinghua University, Beijing, 100084, China. (mengqi-lei@163.com, lisiqi19971013@gmail.com, \{dinggg, gaoyue\}@tsinghua.edu.cn)
\IEEEcompsocthanksitem Yihong Wu is with the Department of Mechanical Engineering, Taiyuan University of Technology, Taiyuan, 030024, China. (wuyihong0453@link.tyut.edu.cn)
\IEEEcompsocthanksitem Han Hu is with the School of Information and Electronics, Beijing Institute of Technology, Beijing, 100811, China. (hhu@bit.edu.cn)
\IEEEcompsocthanksitem You Zhou and Zongze Wu are with the College of Mechatronics and Control Engineering, and the Guangdong Laboratory of Artificial Intelligence and Digital Economy (SZ), Shenzhen University, Shenzhen, 518060, China. (2450097001@mails.szu.edu.cn, zzwu@szu.edu.cn)
\IEEEcompsocthanksitem Xinhu Zheng is with the Internet of Things Thrust and the Intelligent Transportation Thrust, The Hong Kong University of Science and Technology (Guangzhou), Guangzhou, China, and the Department of Electronic and Computer Engineering, The Hong Kong University of Science and Technology, Hong Kong SAR, China. (xinhuzheng@hkust-gz.edu.cn)
\IEEEcompsocthanksitem Shaoyi Du is with the State Key Laboratory of Human-Machine Hybrid Augmented Intelligence, National Engineering Research Center for Visual Information and Applications, and Institute of Artificial Intelligence and Robotics, Xi'an Jiaotong University, Xi'an, 710049, China. (dushaoyi@xjtu.edu.cn)
\IEEEcompsocthanksitem Mengqi Lei and Siqi Li contributed equally to this work.
}
}

\markboth{YOLOv13: Real-Time Object Detection with Hypergraph-Enhanced Adaptive Visual Perception}%
{Lei \MakeLowercase{\textit{et al.}}: YOLOv13: Real-Time Object Detection with Hypergraph-Enhanced Adaptive Visual Perception}


\maketitle

\input{sections/1_abstract}
\input{sections/2_introduction}

\input{sections/3_related_work}
\input{sections/4_methodology}

\input{sections/5_experiment}

\input{sections/6_conclusion}

 
\bibliographystyle{IEEEtran}
\bibliography{a_cite}
%


\begin{IEEEbiography}[{\includegraphics[width=1in,height=1.25in,clip,keepaspectratio]{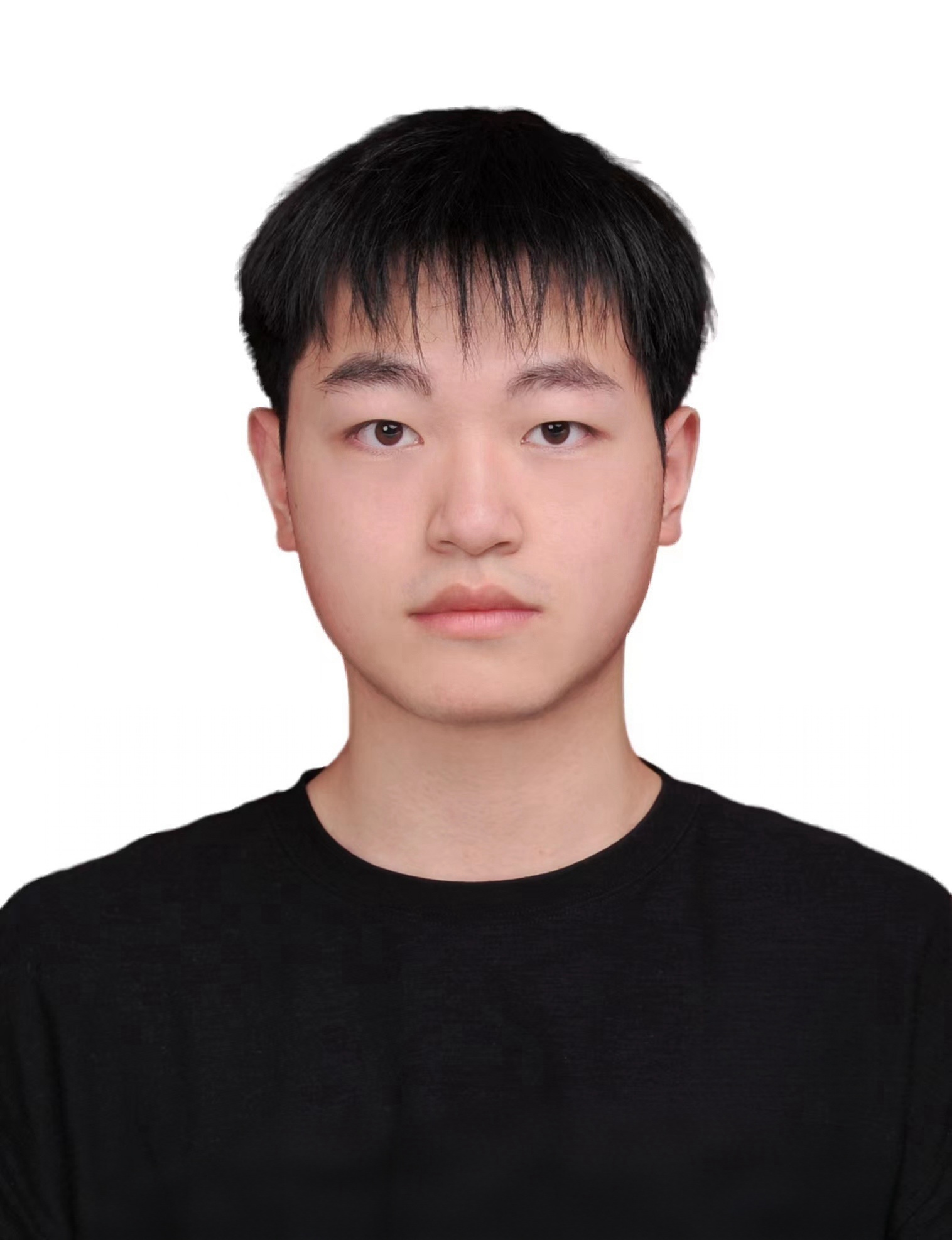}}]{Mengqi Lei} is currently working toward the Ph.D. degree in the School of Software, Tsinghua University, Beijing. He received the B.S. degree from the Department of Computer Science, China University of Geosciences, Wuhan, China. His research interests include hypergraph computation, computer vision, and vision language models.
\end{IEEEbiography}

\begin{IEEEbiography}[{\includegraphics[width=1in,height=1.25in,clip,keepaspectratio]{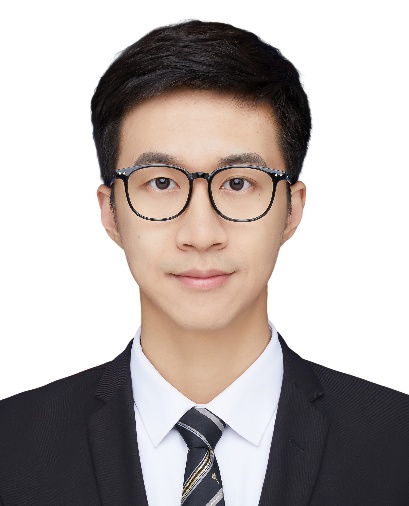}}]{Siqi Li} is a postdoctoral researcher with the School of Software, Tsinghua University. He received the B.S. degree from the Beihang University, Beijing, China, and the Ph.D. degree from Tsinghua University, Beijing, China. His research interests include computer vision and machine learning.
\end{IEEEbiography}

\begin{IEEEbiography}[{\includegraphics[width=1in,height=1.25in,clip,keepaspectratio]{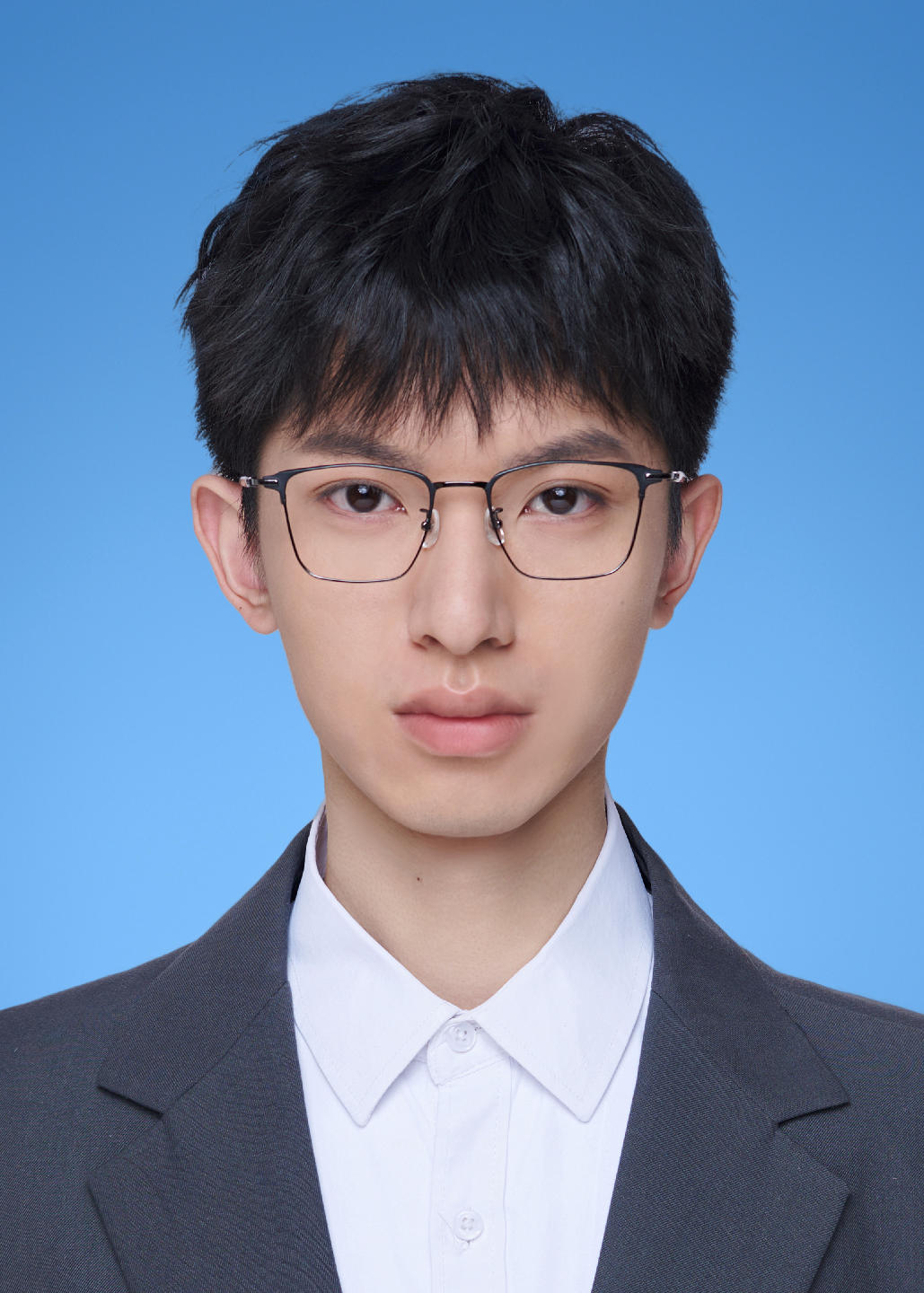}}]{Yihong Wu} 
is an undergraduate student with the College of Mechanical Engineering, Taiyuan University of Technology, Taiyuan, China. He is currently pursuing the B.S. degree in Robotics Engineering. His research interests include computer vision and hypergraph learning.
\end{IEEEbiography}

\begin{IEEEbiography}[{\includegraphics[width=1in,height=1.25in,clip,keepaspectratio]{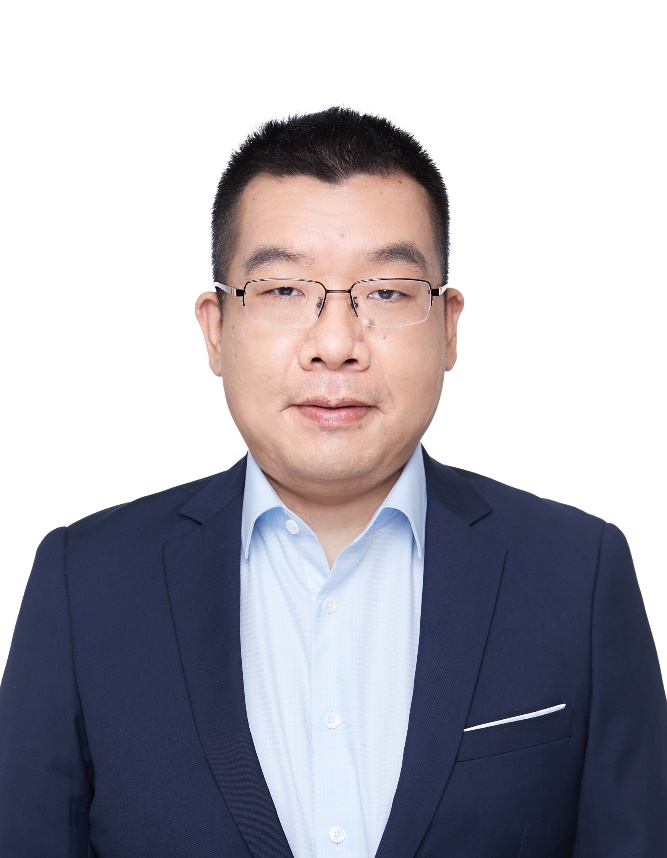}}]{Han Hu} received the B.E. and Ph.D. degrees from the University of Science and Technology of China, Hefei, China, in 2007 and 2012, respectively. He is currently a Professor with the School of Information and Electronics, Beijing Institute of Technology, Beijing, China. His research interests include multimedia networking, edge intelligence, and space–air–ground integrated network. Dr. Hu has served as a TPC Member of Infocom, the ACM International Conference on Multimedia (ACM MM), AAAI, and the International Joint Conference on Artificial Intelligence (IJCAI). He received several academic awards, including the Best Paper Award of the IEEE Transactions on Circuits and Systems for Video Technology (TCSVT) 2019, the Best Paper Award of the IEEE Multimedia Magazine 2015, and the Best Paper Award of the IEEE Globecom 2013. He has served as an Associate Editor for the IEEE Transactions on Multimedia (TMM) and Ad Hoc Networks.
\end{IEEEbiography}

\begin{IEEEbiography}[{\includegraphics[width=1in,height=1.25in,clip,keepaspectratio]{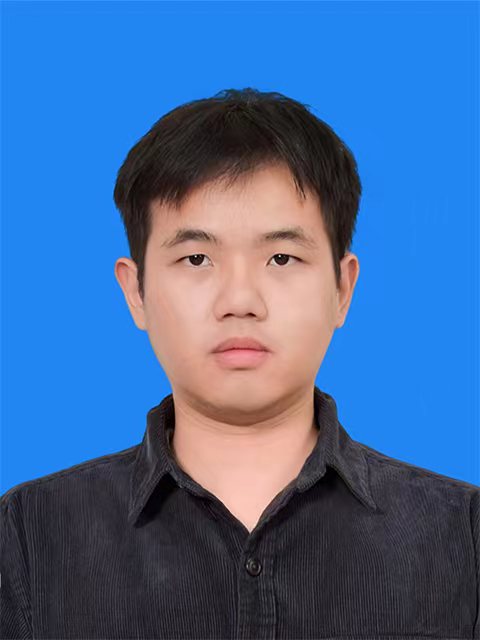}}]{You Zhou} received the M.S. degree from Guangdong University of Technology, Guangzhou, China, in 2024. He is currently pursuing the Ph.D. degree with the College of Mechatronics and Control Engineering, Shenzhen University, Shenzhen, China. His current research interests include machine learning, computer versions, and defect detection.
\end{IEEEbiography}

\begin{IEEEbiography}[{\includegraphics[width=1in,height=1.25in,clip,keepaspectratio]{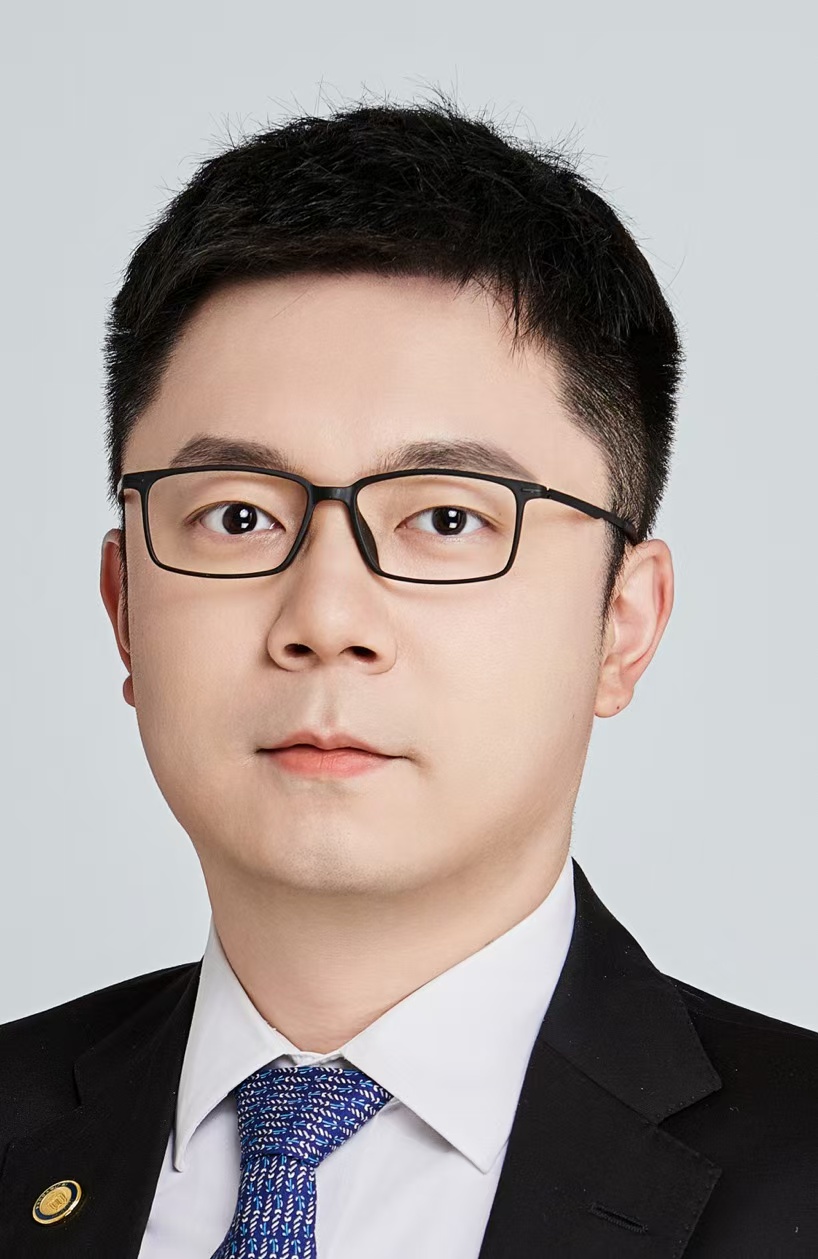}}]{Xinhu Zheng} is currently an Assistant Professor with the Intelligent Transportation Thrust of the Systems Hub, at Hong Kong University of Science and Technology (GZ). He received the Ph.D. degree in Electrical and Computer Engineering from the University of Minnesota, Minneapolis. He has published more than 30 papers on peer-review journals and conferences, including IEEE Internet of Things Journal, IEEE Transactions on Intelligent Transportation Systems, IEEE Transactions on Systems, Man, and Cybernetics: Systems, etc. He is currently an Associate Editor for IEEE Transactions on Intelligent Vehicles. His current research interests include data mining, multi-agent information fusion, multi-modal data fusion and data analysis in intelligent transportation system and ITS related intelligent systems, by exploiting different data modalities, and leveraging various optimization and machine learning techniques.
\end{IEEEbiography}

\begin{IEEEbiography}[{\includegraphics[width=1in,height=1.25in,clip,keepaspectratio]{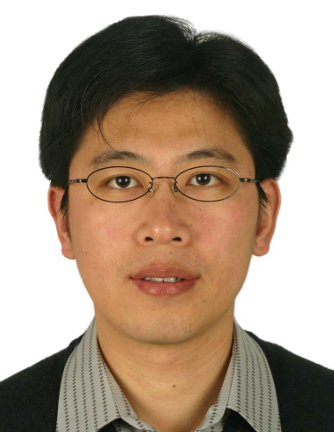}}]{Guiguang Ding} is currently a Distinguished Researcher with the School of Software, Tsinghua University; a Ph.D. Supervisor; an Associate Dean of the School of Software, Tsinghua University; and the Deputy Director of the National Research Center for Information Science and Technology. His research interests mainly focus on visual perception, theory and method of efficient retrieval and weak supervised learning, neural network compression of vision task under edge computing and power limited scenes, visual computing systems, and platform developing. He was the Winner of the National Science Fund for Distinguished Young Scholars.
\end{IEEEbiography}

\begin{IEEEbiography}[{\includegraphics[width=1in,height=1.25in,clip,keepaspectratio]{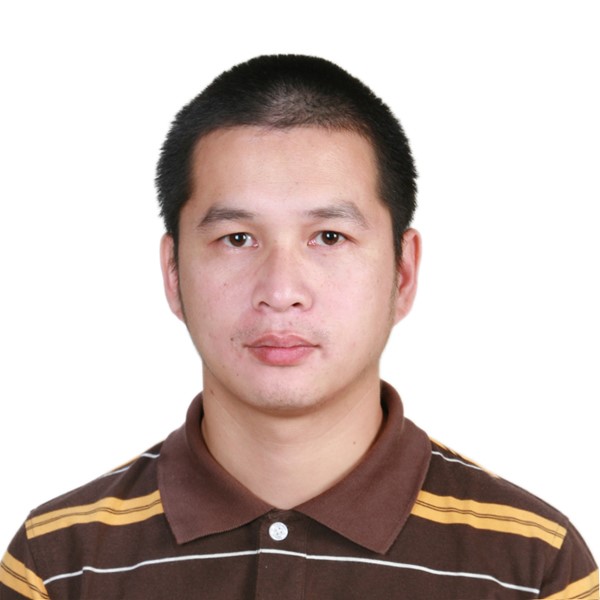}}]{Shaoyi Du} is a professor at Xi’an Jiaotong University. He received double Bachelor degrees in computational mathematics and in computer science in 2002 and received his M.S. degree in applied mathematics in 2005 and Ph.D. degree in pattern recognition and intelligence system from Xi’an Jiaotong University, China in 2009. His research interests include computer vision, machine learning and pattern recognition.
\end{IEEEbiography}

\begin{IEEEbiography}[{\includegraphics[width=1in,height=1.25in,clip,keepaspectratio]{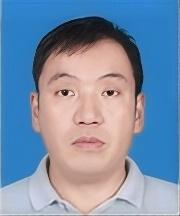}}]{Zongze Wu} is a professor with the College of Mechatronics and Control Engineering, Shenzhen University, and also with the Guangdong Laboratory of Artificial Intelligence and Digital Economy (SZ), Shenzhen, China. He received the PhD degree in pattern reorganization and intelligence system from Xi'an Jiaotong University, in 2005, Xi'an, China.
\end{IEEEbiography}

\begin{IEEEbiography}[{\includegraphics[width=1in,height=1.25in,clip,keepaspectratio]{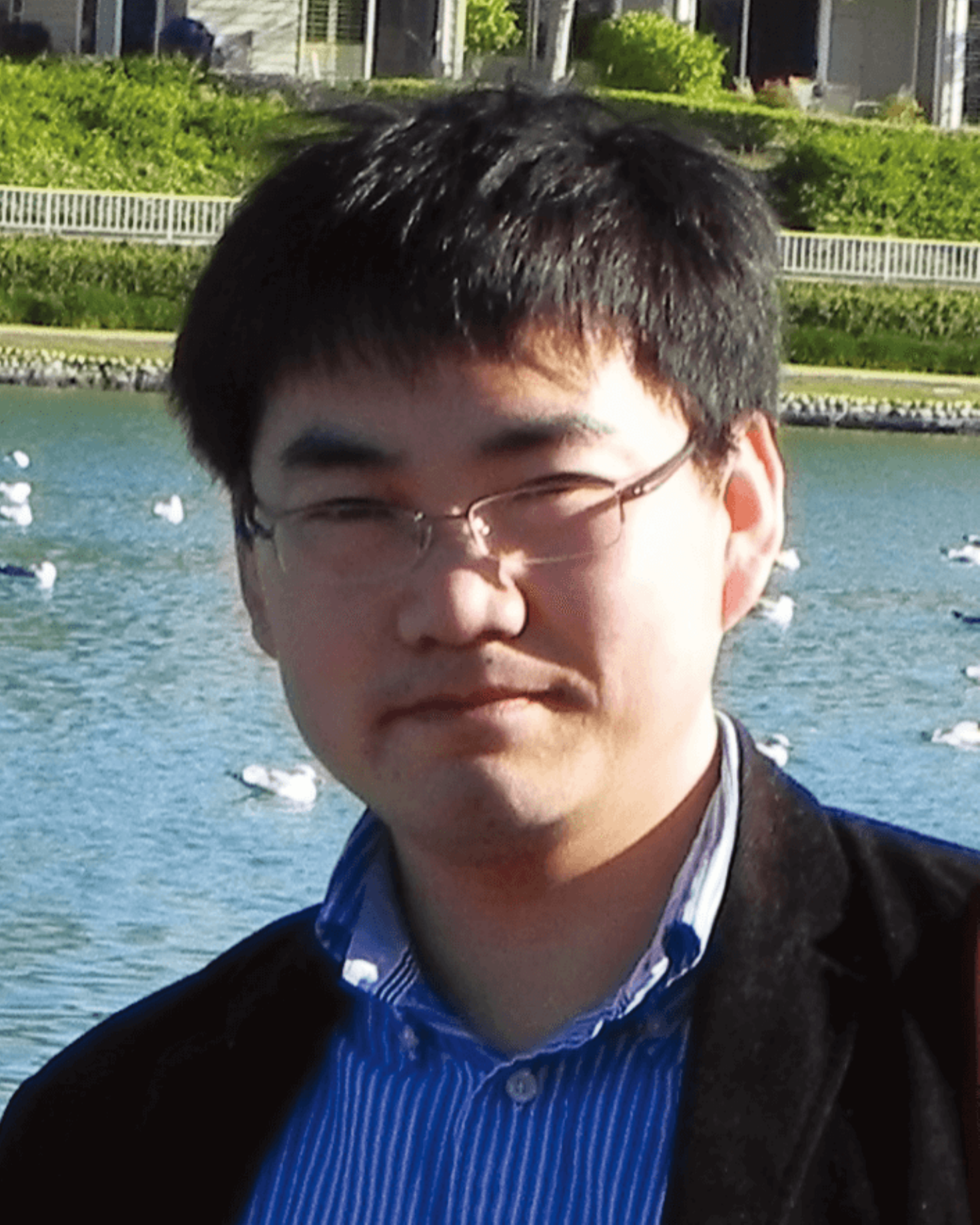}}]{Yue Gao} is an associate professor with the School of Software, Tsinghua University. He received the B.S. degree from the Harbin Institute of Technology, Harbin, China, and the M.E. and Ph.D. degrees from Tsinghua University, Beijing, China.
\end{IEEEbiography}

\vfill

\end{document}

%% file: sections/1_abstract.tex
\begin{abstract}
The YOLO series models reign supreme in real-time object detection due to their superior accuracy and computational efficiency. However, both the convolutional architectures of YOLO11 and earlier versions and the area-based self-attention mechanism introduced in YOLOv12 are limited to local information aggregation and pairwise correlation modeling, lacking the capability to capture global multi-to-multi high-order correlations, which limits detection performance in complex scenarios. In this paper, we propose YOLOv13, an accurate and lightweight object detector. To address the above-mentioned challenges, we propose a Hypergraph-based Adaptive Correlation Enhancement (HyperACE) mechanism that adaptively exploits latent high-order correlations and overcomes the limitation of previous methods that are restricted to pairwise correlation modeling based on hypergraph computation, achieving efficient global cross-location and cross-scale feature fusion and enhancement. Subsequently, we propose a Full-Pipeline Aggregation-and-Distribution (FullPAD) paradigm based on HyperACE, which effectively achieves fine-grained information flow and representation synergy within the entire network by distributing correlation-enhanced features to the full pipeline. Finally, we propose to leverage depthwise separable convolutions to replace vanilla large-kernel convolutions, and design a series of blocks that significantly reduce parameters and computational complexity without sacrificing performance. We conduct extensive experiments on the widely used MS COCO benchmark, and the experimental results demonstrate that our method achieves state-of-the-art performance with fewer parameters and FLOPs. Specifically, our YOLOv13‑N improves mAP by 3.0\% over YOLO11‑N and by 1.5\% over YOLOv12‑N. The code and models of our YOLOv13 model are available at: \texttt{\url{https://github.com/iMoonLab/yolov13}}.

\end{abstract}


%% file: sections/2_introduction.tex
\section{Introduction}

\IEEEPARstart{R}{eal-time} object detection has long been at the forefront of computer vision research~\cite{obj_det_survey,survey_20y}, aiming to localize and classify objects within images with minimal latency, which is critical for a wide range of applications, \eg, industrial anomaly detection, autonomous driving, and video surveillance~\cite{yolo_based_survey}. Recent years have witnessed the dominance of single‑stage CNN detectors that integrate region proposal, classification, and regression into a unified end‑to‑end framework in this field~\cite{rcnn,maskrcnn,yolov1,rt_detr}. Among them, YOLO (You Only Look Once) series~\cite{yolov1,yolov2,yolov3,yolov4,yolov5,yolov6,yolov7,yolov8,yolov9,yolov10,yolo11,yolov12} has become mainstream due to its excellent balance between inference speed and accuracy.

\input{sections/figs/fig1}
From the early YOLO versions to the recent YOLO11 model, architectures centered on convolution have been adopted, aiming to extract image features and achieve object detection through differently designed convolutional layers. The latest YOLOv12 model~\cite{yolov12} further leverages an area-based self-attention mechanism to enhance the representation capability of the model. On the one hand, convolutional operations inherently perform local information aggregation within a fixed receptive field. Thus, the modeling capacity is constrained by the kernel size and network depth. On the other hand, although the self-attention mechanism extends the receptive field, its high computational cost necessitates the use of local area-based computation as a trade-off, thereby preventing adequate global perception and modeling. In addition, the self-attention mechanism can be regarded as the modeling of pairwise pixel correlations on a fully-connected semantic graph, which inherently limits its capacity to capturing only binary correlations and prevents it from representing and aggregating multi-to-multi high-order correlations. Therefore, the architecture of existing YOLO models restricts their capability to model global high-order semantic correlations, leading to performance bottlenecks in complex scenarios. Hypergraphs can model multi-to-multi high-order correlations. Different from traditional graphs, each hyperedge in a hypergraph connects multiple vertices, thus enabling the modeling of correlations between multiple vertices. Some studies~\cite{vihgnn,hgvit,hgformer} have demonstrated the necessity and validity of using hypergraph to model multi-pixel high-order correlations for visual tasks including object detection. However, existing methods simply use manually set threshold parameter values to determine whether pixels are correlated based on pixel feature distances, \ie, pixels with feature distances below a specific threshold are deemed to be correlated. This manual modeling paradigm makes it difficult to cope with complex scenes and leads to additional redundant modeling, resulting in limited detection accuracy and robustness.

To address the above-mentioned challenges, we propose \textbf{YOLOv13}, a novel, real-time, breakthrough end-to-end object detector. Our proposed YOLOv13 model extends traditional region-based pairwise interaction modeling to global high-order correlation modeling, enabling the network to perceive deep semantic correlations across spatial positions and scales, which significantly enhances detection performance in complex scenarios. Specifically, to overcome the limitations in robustness and generalization caused by the handcrafted hyperedge construction in existing methods, we propose a novel \textbf{Hyper}graph-based \textbf{A}daptive \textbf{C}orrelation \textbf{E}nhancement mechanism, named \textbf{HyperACE}. HyperACE takes pixels in multi-scale feature maps as vertices and adopts a learnable hyperedge construction module to adaptively explore high-order correlations between vertices. Then, a message-passing module with linear complexity is leveraged to effectively aggregate multi-scale features with the guidance of high-order correlations to achieve effective visual perception of complex scenarios. In addition, low-order correlation modeling is also integrated in HyperACE for complete visual perception. Building on HyperACE, we propose a novel YOLO architecture containing a \textbf{Full}‑\textbf{P}ipeline \textbf{A}ggregation‑and-\textbf{D}istribution paradigm, named \textbf{FullPAD}. Our proposed FullPAD aggregates multi-level features extracted from the backbone network using the HyperACE mechanism, and then distributes the correlation-enhanced features to the backbone, neck, and detection head to achieve fine‑grained information flow and representational synergy across the entire pipeline, which significantly improves the gradient propagation and enhances the detection performance. Finally, to reduce model size and computational cost without sacrificing performance, we propose a series of lightweight feature extraction blocks based on depthwise separable convolutions. By replacing large‑kernel vanilla convolutional blocks with the depthwise separable convolution blocks, faster inference speed and reduced model size can be achieved, resulting in a better trade-off between efficiency and performance. To validate the effectiveness and efficiency of our proposed model, we conduct extensive experiments on the widely used MS COCO benchmark~\cite{mscoco}. The quantitative and qualitative experimental results demonstrate that our proposed method outperforms all previous YOLO models and variants while remaining lightweight. In particular, YOLOv13‑N/S achieve 1.5\%/0.9\% and 3.0\%/2.2\% mAP improvements compared to YOLOv12‑N/S and YOLO11‑N/S, respectively. Ablation experiments further demonstrate the effectiveness of each proposed module.

Our contributions are summarized as follows:
\begin{itemize}
    \item We propose YOLOv13, a superior real-time end-to-end object detector. Our YOLOv13 model uses adaptive hypergraphs to explore potential high-order correlations, and achieves accurate and robust detection based on effective information aggregation and distribution with the guidance of high-order correlations.
    \item We propose HyperACE mechanism to capture the latent high-order correlations in complex scenes based on adaptive hypergraph computation and achieve feature enhancement based on correlation guidance. We propose a FullPAD paradigm to enable multi‑scale feature aggregation and distribution within the entire pipeline, enhancing information flow and representational synergy. We propose a series of lightweight blocks based on depthwise separable convolutions to replace large-kernel vanilla convolutional blocks, significantly reducing the number of parameters and computational complexity.
    \item We conduct extensive experiments on the MS COCO benchmark. The experimental results demonstrate that our YOLOv13 achieves state-of-the-art detection performance while maintaining lightweight.
\end{itemize}

%% file: sections/figs/fig1.tex
\begin{figure}[!tbp]
    \centering
    \includegraphics[width=1\linewidth]{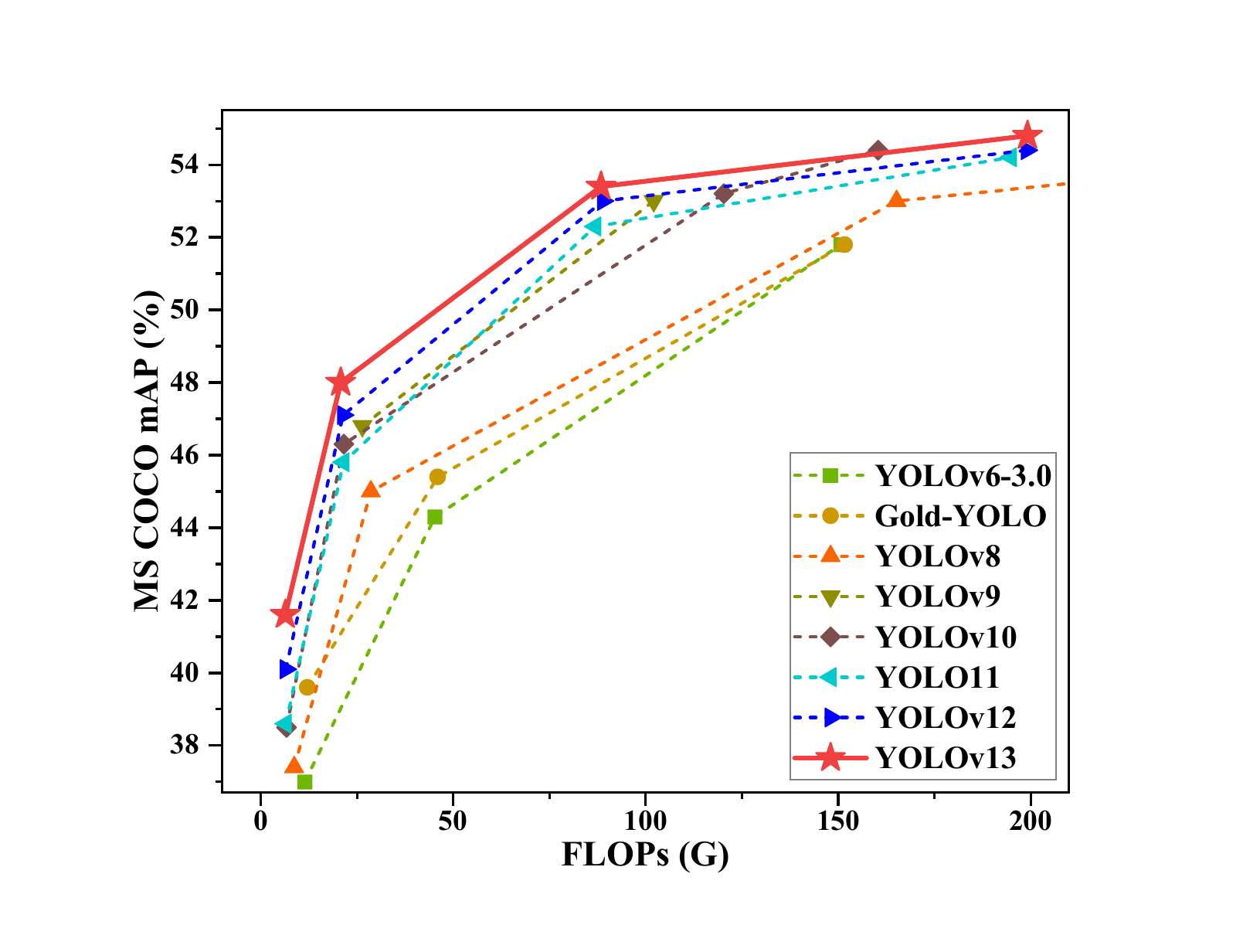}
    \vspace{-0.6cm}
    \caption{Comparison of our proposed YOLOv13 model with previous YOLO models on the MS COCO dataset. Our proposed model can achieve higher detection accuracy with lower computational complexity.}
     \vspace{-0.6cm}
    \label{fig:fig1}
\end{figure}

%% file: sections/3_related_work.tex
\section{Related Works}
\subsection{Evolution of YOLO Detectors}
Since the advent of deep convolutional neural networks~\cite{survey_20y}, real‑time object detection has rapidly evolved from multi‑stage pipelines exemplified by the R‑CNN series~\cite{rcnn,fast_rcnn,faster_rcnn, maskrcnn} to highly optimized single‑stage frameworks epitomized by the YOLO models~\cite{yolo_survey}. The original YOLO (You Only Look Once) model~\cite{yolov1} first recast detection as a single‑shot regression problem, eliminating proposal‑generation overhead and achieving an excellent speed-accuracy trade‑off. Subsequent YOLO iterations have continually refined architecture and training strategies~\cite{yolo_based_survey,yolo_survey}. YOLOv2~\cite{yolov2} boosted precision by introducing anchor‑based predictions and a DarkNet‑19 backbone. YOLOv3~\cite{yolov3} strengthened small‑object detection with a DarkNet‑53 backbone and three‑scale predictions. YOLOv4 through YOLOv8~\cite{yolov4,yolov5,yolov6,yolov7,yolov8} progressively integrated modules such as CSP, SPP, PANet, multi‑mode support, and anchor‑free heads to further balance throughput and accuracy. YOLOv9~\cite{yolov9} and YOLOv10~\cite{yolov10} then focused on lightweight backbones and streamlined end‑to‑end deployment. Later, YOLO11~\cite{yolo11} preserved the ``backbone-neck-head'' modular design but replaced the original C2f block with the more efficient C3k2 unit and added a Convolutional Block with Partial Spatial Attention (C2PSA) to enhance detection of small and occluded targets. The latest YOLOv12~\cite{yolov12} marks the full integration of attention mechanisms, introducing the Residual Efficient Layer Aggregation Network (R‑ELAN) alongside lightweight Area Attention (A2) and Flash Attention to optimize memory access, thereby achieving efficient global and local semantic modeling while maintaining real‑time performance and improving robustness and precision.

Meanwhile, some YOLO‑based variants have emerged~\cite{yolo_survey}. YOLOR~\cite{yolor} fused explicit and implicit features for richer representations and stronger generalization. YOLOX~\cite{yolox} employed an anchor‑free head with dynamic label assignment to simplify the pipeline and improve small‑object detection. YOLO‑NAS~\cite{yolonas} leveraged AutoNAC for neural architecture search, using Quant‑Aware RepVGG and mixed‑precision quantization to optimize throughput and small‑object performance. Gold‑YOLO~\cite{goldyolo} introduced a GD mechanism to enhance multi‑scale feature fusion capabilities. YOLO-MS~\cite{yoloms} introduced an MS-Block with integrated Global Query Learning for dynamic multi-branch feature fusion and a progressive Heterogeneous Kernel Size Selection strategy to enrich multi-scale representations with minimal overhead.

However, as mentioned above, the architecture of the current YOLO series models limits them solely to modeling local pairwise correlations, preventing them from modeling global multi-to-multi high-order correlations. This restricts the detection performance of existing methods in complex scenarios.

\input{sections/figs/fig3-pipeline}

\subsection{High-Order Correlation Modeling}
Complex multi-to-multi high-order correlations are ubiquitous in nature, \eg, neural connections and protein interactions, as well as in the field of information science, \eg, social networks~\cite{hg_book, hg_learning}. In visual data, different objects engage in spatial, temporal, and semantic interactions forming complex correlations. These correlations may be pairwise (low‑order) or more complex group‑based correlations (high‑order).
Hypergraphs, the extension of vanilla graphs, can represent not only pairwise correlations but also multi‑to-multi high‑order correlations~\cite{hg_survey,hg_survey2}. In recent years, hypergraph neural networks (HGNNs) have emerged as a primary tool for modeling such high‑order correlations~\cite{hg_derain,hg_medical,hg_network,hyperyolo}. Feng~\etal~\cite{hgnn} proposed spectral‑domain HGNNs, demonstrating the advantage in the visual retrieval task. Gao~\etal~\cite{hgnnp} further proposed HGNN$^+$ with spatial hypergraph convolution operators, enhancing HGNN applicability.
Recently, Feng~\etal\cite{hyperyolo} pioneered the integration of HGNN into detection models, demonstrating the necessity of high-order correlation modeling for detection. However, this method simply uses a handcrafted fixed parameter as threshold, and pixels with feature distances smaller than the threshold are determined to be correlative, leading to inadequate correlation modeling accuracy and robustness.

To address the above-mentioned challenges, we propose a hypergraph-based adaptive correlation enhancement mechanism, which efficiently models cross-location and cross-scale semantic interactions by adaptively exploiting latent correlations. This mechanism overcomes the lack of robustness in existing hypergraph computing paradigms caused by handcrafted hyperparameters, as well as the absence of global high-order correlation modeling in existing YOLO series models.

%% file: sections/figs/fig3-pipeline.tex
\begin{figure*}[!tp]
    \centering
    \includegraphics[width=0.9\linewidth]{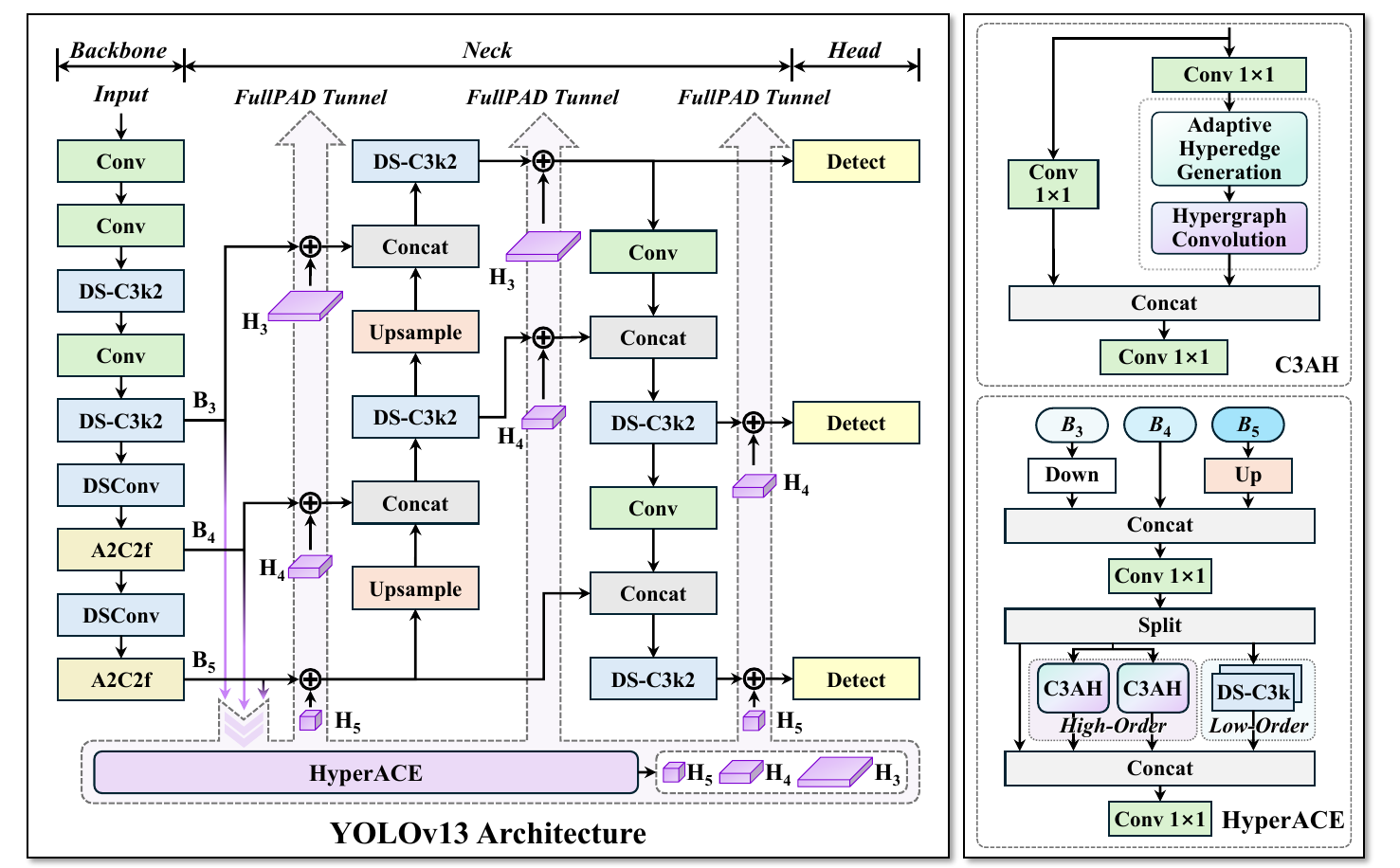}
    \vspace{-0.2cm}
    \caption{The network architecture of our proposed YOLOv13 model. Taking multi-scale features extracted from the backbone as input, HyperACE adaptively explores high-order correlations and achieves feature enhancement and fusion. Then, the correlation-enhanced features are distributed to the entire network by FullPAD tunnels to achieve accurate object detection in complex scenes. The detailed structure of HyperACE is shown on the right.
    }
    \vspace{-0.2cm}
    \label{fig:framework}
\end{figure*}

%% file: sections/4_methodology.tex
\section{Method}
In this section, we introduce our proposed YOLOv13 method. In Sec.~\ref{sec:overall}, we introduce the overall network architecture of our proposed model. Then, the detailed ideology and structure of our proposed hypergraph-based adaptive correlation enhancement mechanism and the full-pipeline aggregation-and-distribution paradigm are presented in Sec.~\ref{sec:HyperACE} and Sec.~\ref{sec:fullpad}, respectively. Finally, in Sec.~\ref{sec:dsc3k}, we introduce the architecture of our proposed lightweight feature extraction blocks.

\subsection{Overall Architecture}
\label{sec:overall}
Previous YOLO series follow the ``Backbone $\to$ Neck $\to$ Head" computational paradigm, which inherently constrains the adequate delivery of information flow. In contrast, our model enhances the traditional YOLO architecture by achieving a \textbf{Full}-\textbf{P}ipeline feature \textbf{A}ggregation-and-\textbf{D}istribution (\textbf{FullPAD}) through the \textbf{Hyper}graph-based \textbf{A}daptive \textbf{C}orrelation \textbf{E}nhancement (\textbf{HyperACE}) mechanism. Thus, our proposed method achieves fine‑grained information flow and representation synergy throughout the network, which can improve gradient propagation and significantly enhance detection performance. Specifically, as shown in Fig.~\ref{fig:framework}, our YOLOv13 model first extracts multi-scale feature maps $B_1, B_2, B_3, B_4, B_5$ using a backbone network similar to previous works, but in which the large-kernel convolutions are replaced with our proposed lightweight DS-C3k2 blocks. Then, different from traditional YOLO methods that directly input $B_3$, $B_4$, and $B_5$ into the neck network, our method gathers and forwards these features into the proposed HyperACE module, achieving cross-scale cross-position features high-order correlation adaptive modeling and feature enhancement. Subsequently, our FullPAD paradigm leverages three separate tunnels to distribute the correlation-enhanced feature to the connection between the backbone and neck, the internal layers of the neck, and the connection between the neck and head, respectively, for better information flow. Finally, the output feature maps of the neck are forwarded into detection heads to achieve multi‑scale object detection. 

\subsection{Hypergraph-Based Adaptive Correlation Enhancement}
\label{sec:HyperACE}

To achieve efficient and robust cross‐scale cross‐position correlation modeling and feature enhancement, we propose a hypergraph-based adaptive correlation enhancement mechanism. As shown in Fig.~\ref{fig:framework}, HyperACE contains two core components, \ie, the global high‐order perception branch based on the C3AH module, in which high‐order visual correlations are modeled with linear complexity using adaptive hypergraph computation, and the local low‐order perception branch based on the DS‐C3k block. In the following subsections, we will introduce the adaptive hypergraph computation, the C3AH module, and the overall design of the HyperACE, respectively.

\subsubsection{Adaptive Hypergraph Computation}
To effectively and efficiently model the high-order correlation in visual features and achieve correlation-guided feature aggregation and enhancement, we propose a novel adaptive hypergraph computation paradigm. Different from traditional hypergraph modeling methods that use manual pre-defined parameters to construct hyperedges based on feature similarity, our proposed method adaptively learns the participation degree of each vertex for each hyperedge, making this computational paradigm more robust and efficient. The traditional hypergraph computing paradigm is more applicable to non-Euclidean data, \eg, social networks, since it contains explicit connection relationships, while our adaptive hypergraph computing paradigm is more conducive to computer vision tasks. 

Specifically, our proposed adaptive hypergraph is defined as $\mathcal{G}= \{\mathcal{V}, \mathcal{A}\}$, where $\mathcal{V}$ is the vertex set, and $\mathcal{A}$ is the adaptive hyperedge set. In an adaptive hypergraph, each hyperedge connects all vertices, where each vertex participates in the hyperedge with a continuous and differentiable contribution degree. Therefore, different from the traditional hypergraph which uses an incidence matrix $H \in \{0,1\}^{N \times M}$ for representation, our adaptive hypergraph is represented using a continuous participation matrix $A \in [0,1]^{N \times M}$, where $N$ and $M$ are the number of vertices and hyperedges, respectively. $A_{i,m}$ denotes the participation degree of vertex $i$ in hyperedge $m$. Accordingly, our adaptive hypergraph computation paradigm consists of two stages, \ie, adaptive hyperedge generation and hypergraph convolution, as shown in Fig.~ \ref{fig:hyperace}.
\input{sections/figs/fig4-HyperACE}

\textbf{Adaptive Hyperedge Generation}.
The adaptive hyperedge generation stage focuses on dynamically modeling correlations from input visual features to generate hyperedges and estimate the participation degree of each vertex to each hyperedge. Specifically, let $X=\bigl\{x_i \in \mathbb{R}^C \mid i = 1, \dots, N\bigr\}$ denote the vertex features, where $C$ is the number of feature channels. Our proposed method first uses global average pooling and max pooling to generate context vectors $f_{\mathrm{avg}}$ and $f_{\mathrm{max}}$, respectively. Then, the context vectors are concatenated to obtain global vertex context $f_{\mathrm{ctx}}$, \ie, 
\begin{equation}
    f_{\mathrm{ctx}}
= \begin{pmatrix} f_{\mathrm{avg}} \\[4pt] f_{\mathrm{max}} \end{pmatrix}
\in \mathbb{R}^{2C}.
\end{equation}
Subsequently, a mapping layer $\phi: \mathbb{R}^{2C} \to \mathbb{R}^{M \times C}$ is leveraged to generate the global offset $\Delta P$ from the vertex context, \ie, $\Delta P = \phi(f_{\mathrm{ctx}})$, where $M$ is the number of hyperedges. The global offset is then added with a learnable global prototype \(P_0 \in \mathbb{R}^{M \times C}\) to obtain $M$ dynamic hyperedge prototypes $P$, \ie, $P = P_0 + \Delta P$. These prototypes represent potential visual correlations within the scene. To calculate the participation degree of each vertex, another projection layer is leveraged to generate the vertex query vector $z_i$ from vertex feature $x_i$, \ie,
\begin{equation}
    z_i = W_{\mathrm{pre}}\,x_i \in \mathbb{R}^C,
\end{equation}
where \(W_{\mathrm{pre}}\) is the weight matrix. 

To further increase feature diversity, we introduce a multi‐head mechanism to split \(z_i\) into \(h\) subspaces $\{ \hat z_i^\tau \in \mathbb{R}^{d_h}\}_{\tau=1}^h$ along the feature dimension, where $d_h = C/h$, and we similarly split each hyperedge prototype into \(h\) subspaces $\{\hat p_m^\tau \in \mathbb{R}^{d_h}\}_{\tau=1}^h$. Then, in the \(\tau\)‐th subspace, the similarity between the $i$-th vertex query vector and the $m$-th prototype is computed:
\begin{equation}
    s_{i,m}^\tau = \frac{\langle \hat z_i^\tau, \hat p_m^\tau\rangle}{\sqrt{d_h}}.
\end{equation}
Thus, the overall similarity is defined as the average of all subspace similarities, \ie, $\bar s_{i,m} = \frac{1}{h} \sum_{\tau=1}^h s_{i,m}^\tau$.
Finally, we use the similarity between the vertex query vector and the prototype as the contribution of the vertex to the hyperedge, and normalize across vertices to obtain the continuous participation matrix \(A\), formulated as:
\begin{equation}
    A_{i,m}
= \frac{\exp\bigl(\bar s_{i,m}\bigr)}
       {\displaystyle\sum_{j=1}^{N}\exp\bigl(\bar s_{j,m}\bigr)}.
\end{equation}

\textbf{Hypergraph Convolution}.
After the adaptive hyperedges are generated, the hypergraph convolution is conducted to achieve feature aggregation and enhancement. Specifically, in the hypergraph convolution, each hyperedge first collects features from all vertices and applies a linear projection to form the hyperedge feature. Then, the hyperedge features are disseminated back to the vertices to update their representations. Formally, this process can be defined as:
\begin{equation}
    \begin{aligned}
        f_{m} &= \sigma\Bigl(W_{e}\sum_{i=1}^{N} A_{i,m}\,x_{i}\Bigr)\\
        \widetilde x_{i} &= \sigma\Bigl(W_{v}\sum_{m=1}^{M} A_{i,m}\,f_{m}\Bigr)
    \end{aligned},
\end{equation}
where $i=1,\dots,N$ is the vertex index, $m=1,\dots,M$ is the hyperedge index, $W_{e}$ and $W_{v}$ denote the hyperedge and vertex projection weights, respectively, and $\sigma$ is the activation function.

\subsubsection{C3AH for Adaptive High-Order Correlation Modeling}
Based on our proposed adaptive hypergraph computation paradigm, we further propose the C3AH block to efficiently capture high-order semantic interactions. Specifically, as shown in Fig.~\ref{fig:framework}, the C3AH block retains the CSP bottleneck branch-split mechanism while integrating an adaptive hypergraph computation module, enabling global high-order semantic aggregation across spatial positions.

Let the input feature map of C3AH block be $X_{\mathrm{in}}\in\mathbb{R}^{C_{\mathrm{in}}\times H\times W}$. We first apply two $1\times1$ convolutional layers to project it into the same hidden dimension,\ie,
\begin{equation}
    X = \text{Conv}_{1\times1}(X_{\text{in}}),\quad
X_{\text{lateral}} = \text{Conv}_{1\times1}(X_{\text{in}}),
\end{equation}
where $X,\,X_{\mathrm{lateral}}\in\mathbb{R}^{C\times H\times W}$ and $C=\lfloor e\,C_{\mathrm{in}}\rfloor$, $e$ is the bottleneck compression ratio. $X$ and $X_{\text{lateral}}$ are used for adaptive hypergraph computation and lateral connection, respectively.

Then, $X$ is flattened as the vertex features and is sent to the adaptive hypergraph computation module, denoted as AHC, to obtain correlation-enhanced features:
\begin{equation}
    X_{h} = \mathrm{AHC}(X).
\end{equation}
Finally, $X_{h}$ and $X_{\mathrm{lateral}}$ are concatenated along the channel dimension and then fused by a $1\times1$ convolutional layer to obtain the output of C3AH block.

\subsubsection{Structure of HyperACE}
HyperACE takes multi‐scale features as input to achieve efficient visual correlation modeling and feature enhancement. Specifically, HyperACE first takes the feature maps $B_3$, $B_4$, and $B_5$ from the last three stages of the backbone as input, then resizes $B_3$ and $B_5$ to the same spatial size as \(B_4\), and aggregates them via a \(1\times1\) convolutional layer to obtain the fused feature $X_b$. Subsequently, \(X_b\) is split along the channel dimension into three feature groups, denoted as $X_b^h \in \mathbb{R}^{C_h\times H\times W}$, $X_b^l \in \mathbb{R}^{C_l\times H\times W}$, and $X_b^s \in \mathbb{R}^{C_s\times H\times W}$, respectively. $X_b^h$, $X_b^l$, and $X_b^s$ are used for global high‑order correlation modeling, local low‑order correlation modeling, and shortcut connection, respectively.

In the high‑order correlation modeling branch, $X_b^h$ is forwarded into $K$ parallel C3AH blocks to explore various latent high-order correlations and obtain enhanced features, \ie, 
\begin{equation}
    \bigl\{X_h^{(k)} = \text{C3AH}_k(X_b^h)\mid k=1,\dots,K\bigr\}.
\end{equation}
These \(K\) enhanced features are then concatenated along the channel dimension to obtain the output of the high‑order correlation modeling branch $X_h$, \ie,
\begin{equation}
    X_h = \mathrm{Concat}\bigl(X_h^{(1)},\,X_h^{(2)},\,\dots,\,X_h^{(K)}\bigr).
\end{equation}
In the local low‑order correlation modeling branch, we use $L$ stacked DS‑C3k modules to capture fine‑grained local information, \ie, 
\begin{equation}
    X_l
= \text{DS-C3k}\bigl(X_b^l\bigr).
\end{equation}
The shortcut branch directly retains the original visual information, \ie, $X_s = X_b^s$.

Finally, the outputs of three branches are concatenated along the channel dimension and fused by a \(1\times1\) convolutional layer to obtain the final output of HyperACE, formulated as:
\begin{equation}
    Y = \mathrm{Conv}_{1\times1}\!\bigl(\mathrm{Concat}(X_h,\,X_l,\,X_s)\bigr).
\end{equation}

Our proposed HyperACE fully leverages the parallel global high‑order correlation modeling branch and the local low‑order correlation modeling branch, while preserving shortcut information, achieving complementary multi‑level visual correlation perception across global-local and high-low orders.

\subsection{Full-Pipeline Aggregation-and-Distribution Paradigm}
\label{sec:fullpad}
To fully utilize the correlation-enhanced features obtained from HyperACE, we further introduce the FullPAD paradigm. Specifically, FullPAD collects multi‐scale feature maps from the backbone and forwards them into HyperACE, and then redistributes the enhanced features to various locations throughout the pipeline via different FullPAD tunnels, as is shown in Fig.~\ref{fig:framework}. This design enables fine‐grained information flow and representation synergy, significantly improving gradient propagation and enhancing detection performance.

In practice, after the correlation-enhanced feature $Y$ is obtained from $B_3$, $B_4$, $B_5$ using HyperACE, FullPAD resizes $Y$ to the spatial resolution of each stage and adjusts its channel dimension using a \(1\times1\) convolutional layer, \ie, 
\begin{equation}
    H_i = \text{Conv}_{1\times1}\bigl(\text{Resize}(Y \leftarrow \mathrm{size}(B_i))\bigr),
\quad i \in \{3,4,5\}.
\end{equation}
Then, for an arbitrary feature map \(F_i\) at stage \(i\), a gated fusion is leveraged to achieve information flow and fusion:
\begin{equation}
    \widetilde{F_i}= F_i+\gamma H_i,
\end{equation}
where \(\gamma\) is a learnable scalar parameter that adaptively balances the contribution of the correlation‐enhanced feature and original feature. In practice, FullPAD transmits the output correlation-enhanced features to seven different destinations via three different tunnels, as shown in Fig.~\ref{fig:framework}.

Based on our proposed FullPAD paradigm, correlation-enhanced features are effectively integrated into different stages of the entire pipeline, enabling the model to fully utilize the visual correlation information to perceive complex scenarios effectively.

\subsection{Model Lightweighting with Depth-Separable Convolution}
\label{sec:dsc3k}

In our proposed YOLOv13, we leverage large‐kernel depthwise-separable convolution (DSConv) as the basic unit to design a series of lightweight feature extraction blocks, as shown in Fig.~\ref{fig:dsblock}, which significantly reduces the number of parameters and computational complexity without compromising model performance.

\textbf{DSConv.} The DSConv block first applies a standard depthwise separable convolutional layer to extract features, and then leverages batch normalization and SiLU activation to obtain the output, \ie,
\begin{equation}
    X_{\mathrm{DS}}
=\mathrm{SiLU}\bigl(\mathrm{BN}\bigl(\mathrm{DSConv}(X_{\mathrm{in}})\bigr)\bigr).
\end{equation}

\textbf{DS‐Bottleneck.} In the DS‐Bottleneck block, we cascade two DSConv blocks, among which the first block is a fixed $3\times3$ depthwise‐separable convolution, and the second block is a large‐kernel ($k\times k$) depthwise‐separable convolution, \ie,
\begin{equation}
    X_{\mathrm{DS2}}
= \mathrm{DSConv}_{k\times k}\bigl(\mathrm{DSConv}_{3\times3}(X_{\mathrm{in}})\bigr).
\end{equation}
Meanwhile, if the input and output have the same number of channels, a residual skip connection will be added to preserve low‐frequency information.
\input{sections/figs/fig5-dsblock}

\textbf{DS-C3k.} The DS‐C3k block inherits from the standard CSP-C3 structure~\cite{yolov5}. Specifically, the input feature is first forwarded into a $1\times1$ convolutional layer to reduce feature channels, and is then processed by $n$ cascaded DS‐Bottleneck blocks. Meanwhile, a lateral $1\times1$ convolutional branch is applied to the input feature. Finally, the features from two branches are concatenated along the channel dimension, and a $1\times1$ convolutional layer is leveraged to restore the feature channels. This design retains the cross‐channel branching of CSP structure while integrating depthwise‐separable lightweight bottlenecks.

\textbf{DS-C3k2.} The DS‐C3k2 block is derived from the C3k2 structure~\cite{yolo11}. Specifically, a $1\times1$ convolutional layer is first applied to unify the channels. Then, the features are split into two parts, with one part fed into multiple DS-C3k modules and the other passed through a shortcut connection. Finally, the outputs are concatenated and fused with a $1\times1$ convolutional layer.

As shown in Fig~\ref{fig:framework}, our proposed YOLOv13 model extensively uses DS-C3k2 blocks as the basic feature extraction module in both the Backbone and the Neck. In HyperACE, we leverage the DS‐C3k block as the low‐order feature extractor. This design achieves up to 30\% parameter reduction and up to 28\% GFLOPs reduction across all YOLOv13 model sizes.

Using our proposed YOLOv13 model, the latent correlations in visual features are adaptively modeled, and by fully propagating the correlation-enhanced features throughout the full pipeline, accurate and efficient object detection in complex scenes can be achieved.

%% file: sections/figs/fig4-hyperace.tex
\begin{figure}[t]
    \centering
    \includegraphics[width=1\linewidth]{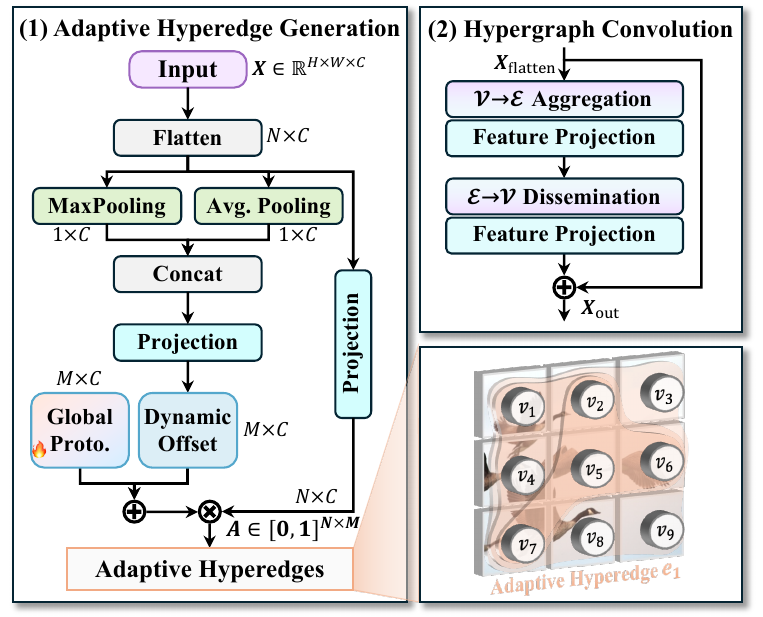}
    \vspace{-0.6cm}
    \caption{Details of the adaptive hypergraph construction and convolution.}
    \vspace{-0.4cm}
    \label{fig:hyperace}
\end{figure}

%% file: sections/figs/fig5-dsblock.tex
\begin{figure}[!tbp]
    \centering
    \includegraphics[width=1\linewidth]{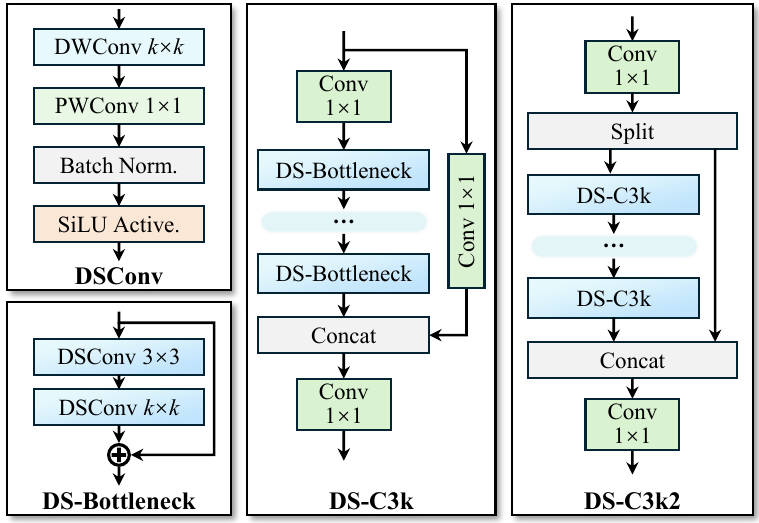}
    \vspace{-0.5cm}
    \caption{Detailed architecture of our proposed DS-series blocks.}
    \vspace{-0.4cm}
    \label{fig:dsblock}
\end{figure}

%% file: sections/5_experiment.tex
\input{sections/tables/coco_compare}

\section{Experiments}
To validate the effectiveness and efficiency of our proposed YOLOv13 model, we conduct extensive experiments. In Sec.~\ref{sec:setting}, we introduce the detailed experimental settings. Then, in Sec.~\ref{sec:cmp}, we compare our proposed method with other existing real-time object detection methods to demonstrate the validity of our method. Finally, we conduct ablation experiments in Sec.~\ref{sec:ablation} to prove the effectiveness of each proposed module.

\subsection{Experimental Setting}
\label{sec:setting}
 
\subsubsection{Dataset}
We use the Microsoft COCO (MS COCO) dataset~\cite{mscoco}, the most widely adopted benchmark for the object detection task, to evaluate our YOLOv13 model and other state‐of‐the‐art real‐time detectors. The MS COCO dataset training set (Train2017) contains approximately 118,000 images, and the validation set (Val2017) contains approximately 5,000 images, covering 80 common object categories in natural scenes. In our implementation, all methods are trained on the Train2017 subset and tested on the Val2017 subset.

As the YOLO series continues to evolve, models with greater generality and stronger generalization capabilities become increasingly important. To this end, we include cross-domain generalization under distribution shifts in our evaluation. As a supplementary benchmark, we select the Pascal VOC 2007 dataset~\cite{voc}, which contains a combined total of 5,011 images in the training and validation sets and 4,952 images in the test set, covering 20 common object classes. To evaluate the cross-domain generalization ability, all methods are directly evaluated on the Pascal VOC 2007 test set using the model trained on the MS COCO dataset.

\subsubsection{Implementation Details}
Similar to previous YOLO models, our YOLOv13 family includes four variants: Nano (N), Small (S), Large (L), and Extra-Large (X). For the N, S, L, and X models, the number of hyperedges $M$ is set to 4, 8, 8, and 12, respectively. For all variants, we train the model for 600 epochs with a batch size of 256. The initial learning rate is 0.01 and SGD is used as the optimizer, which is consistent with YOLO11 and YOLOv12 models. A linear decay scheduler is adopted, and a linear warm‐up is applied within the first 3 epochs. During training, the input image size is set to \(640 \times 640\). We employ the same data augmentation techniques as previous YOLO versions, including Mosaic and Mixup. 
We use 4 and 8 RTX 4090 GPUs to train YOLOv13-N and YOLOv13-S, respectively, and use 4 and 8 A800 GPUs to train YOLOv13-L and YOLOv13-X, respectively.
Additionally, following the standard practice of previous YOLO series, we evaluate the latency on a single Tesla T4 GPU using TensorRT FP16 for all models. 
In addition, it should be noted that, to ensure fair and rigorous comparison, we reproduce all variants of the previous YOLO11 and YOLOv12 (v1.0 version) models using their official settings on the same hardware platform as our YOLOv13 model.

\subsection{Comparison with Other Methods}
\label{sec:cmp}
Table~\ref{tab:coco_compare} shows the quantitative comparison results on the MS COCO dataset. Our proposed method is compared with previous YOLO series models. As mentioned above, our YOLOv13 model and the latest YOLO11 and YOLOv12 models are trained on the same GPUs while the existing methods are trained using their official code and training parameters. From the table, we can observe that all variants of our YOLOv13 model achieve state-of-the-art performance while remaining lightweight. Specifically, compared with the previous YOLOv12 model, our YOLOv13 model can improve $\text{AP}_{50:95}^\text{val}$ by 1.5\%, 0.9\%, 0.4\%, and 0.4\%, and improve $\text{AP}_{50}^\text{val}$ by 1.8\%, 1.0\%, 0.9\%, and 0.9\% in terms of Nano, Small, Large, and Extra-Large models, respectively. In addition, compared with ViT-based methods, our proposed YOLOv13 model can also achieve better detection accuracy with fewer parameters and lower computational complexity. Compared to RT-DETRv2-R18, our YOLOv13-S model can improve $\text{AP}_{50:95}^\text{val}$ by 0.1\% while reducing the number of parameters by 55.0\% and the FLOPs by 65.3\%. Furthermore, from the table, we can observe that our proposed method can achieve more significant advantages in lightweight variants, \eg, the Nano model. This is also the core objective of the YOLO series of models, \ie, more accurate, faster, and lighter. This is because our proposed core HyperACE mechanism can more fully explore the multi-to-multi correlations in complex scenarios. As the high-order version of the traditional self-attention mechanism, HyperACE leverages high-order correlations as guidance to achieve accurate feature enhancement with low parameter counts and computational complexity. These quantitative comparison results prove the effectiveness of our proposed YOLOv13 model.
\input{sections/figs/fig6-box}

\input{sections/tables/voc_comp}

Figure~\ref{fig:vis_bbox} shows the qualitative comparison results on the MS COCO dataset. Our proposed YOLOv13 model is compared with the existing YOLOv10~\cite{yolov10}, YOLO11~\cite{yolo11}, and YOLOv12~\cite{yolov12} models. From the figure, we can observe that our YOLOv13 model can achieve more accurate detection performance in complex scenes. Specifically, as shown in the last row on the left side of Fig.~\ref{fig:vis_bbox}, our YOLOv13-N model can accurately detect objects in complex multi-object scenes. In contrast, previous models miss small objects such as bowls and vases. This is due to the fact that our proposed HyperACE can establish high-order correlations between multiple related objects, enabling accurate detection of multiple targets in complex scenes. As shown in the second row on the right side, only our method successfully detects the plant behind the vase, which is a challenging scenario. Intuitively, the vase and the plant are strongly associated, leading to a high probability appearance of the plant. Our method can achieve accurate detection by mining such latent correlations. As shown in the third row on the right side, our method can accurately detect the tennis racket held by the athlete, while previous methods either missed it or incorrectly detected the shadow. These qualitative results demonstrate the effectiveness of our proposed method.

As mentioned above, to validate the generalization ability of our proposed method, we train our YOLOv13 model and previous YOLO models on the MS COCO dataset and test all methods on the Pascal VOC 2007 dataset. The quantitative results are shown in Tab.~\ref{tab:voc_compare}. From the table, we can observe that our proposed YOLOv13 model achieves satisfactory generalization performance. Specifically, compared to the previous YOLOv12 model, our proposed method improves the $\text{AP}_{50:95}^\text{val}$ by 1.0\% and 0.4\% in terms of Nano and Small models, respectively. More significant performance improvements can be achieved compared to earlier models. These results demonstrate the generalization ability of our proposed method.

\subsection{Ablation Study}
\label{sec:ablation}
\subsubsection{FullPAD and HyperACE} To validate the effectiveness and necessity of our proposed FullPAD paradigm and HyperACE mechanism, we evaluate the performance of the proposed YOLOv13-Small model when FullPAD distributes the features to different locations. The quantitative results are shown in Tab.~\ref{tab:distribution}. Specifically, when FullPAD does not distribute any features, it is equivalent to removing the proposed HyperACE, and the results under such setting are shown in the first row of the table. From the table, we can observe that the removal of the proposed HyperACE will reduce the $\text{AP}_{50:95}^\text{val}$ and $\text{AP}_{50}^\text{val}$ by 0.9\% and 1.1\%, respectively. This result demonstrates the effectiveness of adaptive correlation enhancement. In addition, when FullPAD distributes enhanced features only to the backbone-neck (left FullPAD tunnel in Fig.~\ref{fig:framework}), in-neck (middle FullPAD tunnel in Fig.~\ref{fig:framework}), and neck-head (right FullPAD tunnel in Fig.~\ref{fig:framework}), the $\text{AP}_{50:95}^\text{val}$ will decrease by 0.2\%, 0.4\%, and 0.3\%, respectively, compared to the full model. These results demonstrate the necessity of our proposed FullPAD paradigm.

\input{sections/tables/distribution}
\input{sections/tables/num_edge}
\input{sections/tables/dsblocks}

Figure~\ref{fig:vis_soft} shows the visualization of representative hyperedges generated by adaptive hypergraph construction in the proposed HyperACE module. From the figure, we can intuitively observe the high-order visual correlations captured by the proposed HyperACE, which helps to enhance the interpretability of the model. As shown in the first and second column of the figure, HyperACE can effectively model the correlations among multiple foreground objects or partial elements within a scene, \eg, the skis and skistick in the first row, and the cars and multiple traffic lights in the second row. As shown in the third column, our proposed method can also model the correlations between foreground objects and the background scene, \eg, the tennis racket and tennis court in the third row, and the baseball glove and the baseball field in the last row. These visualizations demonstrate the modeling capabilities of our proposed HyperACE module for potential high-order correlations within scenes, enabling the model to enhance features based on multi-to-multi correlations rather than just low-order pairwise correlations.
\input{sections/figs/fig7-edge}
\input{sections/tables/epoch}
\input{sections/tables/latency}

\subsubsection{Number of hyperedges} To validate the effect of the number of hyperedges $M$ on model performance, we set different numbers of hyperedges and test the performance of the YOLOv13-S model. Table~\ref{tab:num_edge} shows the quantitative results. From the table, we can observe that fewer hyperedges result in fewer model parameters and less computational effort, but also lead to a decline in performance. This is due to insufficient correlation modeling of the scene. When the number of hyperedges is increased to 16, the detection performance still improves, but it also brings additional parameters and computational costs. Similar results can be observed for other model variants. Therefore, we set the number of hyperedges for the N, S, L, and X models to 4, 8, 8, and 12, respectively, to balance performance and computational complexity.

\subsubsection{DS blocks} To demonstrate the effectiveness and efficiency of the proposed DS-series blocks, we validate the performance and FLOPs of replacing the DS-series blocks in our YOLOv13-N and Small models with vanilla convolutions. The quantitative results are shown in Tab.~\ref{tab:dsblocks}. From the table, we can observe that the replacement of vanilla convolution blocks with our proposed DS series blocks leads to only a 0.1\% decrease on  $\text{AP}_{50}^\text{val}$ and no decrease on $\text{AP}_{50:95}^\text{val}$ at all, the FLOPs can be reduced by 1.1 G and 4.2 G, and the number of parameters can be reduced by 0.6 M and 2.2 M for the Nano and Small models, respectively. These results prove the efficiency and validity of our proposed DS series blocks.

\subsubsection{Training epochs} Table~\ref{tab:epoch} shows the effect of training epochs on model performance. We validate the performance of Nano and Small models with different numbers training epochs. From the table, we can observe that the best performance could be achieved when trained for 600 epochs, \ie, 41.6\% and 48.0\% in terms of $\text{AP}_{50:95}^\text{val}$ for YOLOv13 Nano and Small models, respectively. More training epochs will lead to overfitting and performance degradation.

\subsubsection{Latency on different hardware platforms} Table~\ref{tab:latency} shows the inference latency for all variants of our proposed method on different hardware platforms. For the Nano model of our YOLOv13, the inference latency of 1.25 ms and 1.97 ms are achieved on the RTX 4090 and the Tesla T4 GPU, respectively. Considering the deployment condition without a GPU, the Nano model can also achieve an inference speed of 25 FPS (39.97 ms) on a CPU (Intel Xeon Platinum 8352V). For the Small model with better performance, the inference latency on the Tesla T4 GPU is still less than 3 ms. For the Extra-Large model of YOLOv13, the inference latency on the Tesla T4 GPU is 14.67 ms, while on the 4090 GPU it is only 3.1 ms. These results show the efficiency of our YOLOv13.


%% file: sections/tables/coco_compare.tex
\begin{table*}[t]
\centering
\caption{Quantitative comparison with other state-of-the-art real-time object detectors on MS COCO dataset.}
\vspace{-0.2cm}
\label{tab:coco_compare}
\renewcommand{\arraystretch}{1.}
\setlength{\tabcolsep}{5.8mm}
\begin{tabular}{lcccccc}
\toprule
\textbf{Method} & \textbf{FLOPs (G)} & \textbf{Parameters (M)} & \textbf{$\text{AP}_{50:95}^\text{val}$} & \textbf{$\text{AP}_{50}^\text{val}$} & \textbf{$\text{AP}_{75}^\text{val}$} & \textbf{Latency (ms)} \\ \midrule
YOLOv6-3.0-N~\cite{yolov6}        & 11.4  & 4.7  & 37.0  & 52.7  & –    & 2.74\\
Gold-YOLO-N~\cite{goldyolo}        & 12.1  & 5.6  & 39.6  & 55.7  & –    & 2.97\\
YOLOv8-N~\cite{yolov8}            & 8.7   & 3.2  & 37.4  & 52.6  & 40.5 & 1.77\\
YOLOv10-N~\cite{yolov10}           & 6.7& 2.3& 38.5& 53.8& 41.7& 1.84\\
YOLO11-N~\cite{yolo11}            & 6.5   & 2.6  & 38.6  & 54.2  & 41.6 & 1.53\\
YOLOv12-N~\cite{yolov12}           & 6.5   & 2.6  & 40.1  & 56.0  & 43.4 & 1.83\\\rowcolor{blue!10}
\textbf{YOLOv13-N}& \textbf{6.4}   & \textbf{2.5}  & \textbf{41.6} & \textbf{57.8} & \textbf{45.1} & \textbf{1.97}\\ \midrule
YOLOv6-3.0-S~\cite{yolov6}        & 45.3  & 18.5 & 44.3  & 61.2  & –    & 3.42\\
Gold-YOLO-S~\cite{goldyolo} & 46.0  & 21.5 & 45.4  & 62.5  & –    & 3.82\\
YOLOv8-S~\cite{yolov8}            & 28.6  & 11.2 & 45.0  & 61.8  & 48.7 & 2.33\\
RT-DETR-R18~\cite{rt_detr}         & 60.0  & 20.0 & 46.5  & 63.8  & –    & 4.58\\
RT-DETRv2-R18~\cite{rt_detrv2}       & 60.0  & 20.0 & 47.9  & 64.9  & –    & 4.58\\
YOLOv9-S~\cite{yolov9}            & 26.4  & 7.1  & 46.8  & 63.4  & 50.7 & 3.44\\
YOLOv10-S~\cite{yolov10}           & 21.6  & 7.2  & 46.3& 63.0& 50.4& 2.53\\
YOLO11-S~\cite{yolo11}            & 21.5  & 9.4  & 45.8  & 62.6  & 49.8 & 2.56\\
YOLOv12-S~\cite{yolov12}           & 21.4  & 9.3  & 47.1  & 64.2  & 51.0 & 2.82\\ \rowcolor{blue!10}
\textbf{YOLOv13-S}& \textbf{20.8}  & \textbf{9.0}  & \textbf{48.0} & \textbf{65.2} & \textbf{52.0} & \textbf{2.98}\\ \midrule
YOLOv6-3.0-L~\cite{yolov6}        & 150.7 & 59.6 & 51.8  & 69.2  & –    & 9.01\\
Gold-YOLO-L~\cite{goldyolo}         & 151.7 & 75.1 & 51.8  & 68.9  & –    & 10.69\\
YOLOv8-L~\cite{yolov8}            & 165.2 & 43.7 & 53.0  & 69.8  & 57.7 & 8.13\\
RT-DETR-R50~\cite{rt_detr}         & 136.0 & 42.0 & 53.1  & 71.3  & –    & 6.93\\
RT-DETRv2-R50~\cite{rt_detrv2}       & 136.0 & 42.0 & 53.4  & 71.6  & –    & 6.93\\
YOLOv9-C~\cite{yolov9}            & 102.1 & 25.3 & 53.0  & 70.2  & 57.8 & 6.64\\
YOLOv10-L~\cite{yolov10}           & 120.3& 24.4& 53.2& 70.1& 57.2& 7.31\\
YOLO11-L~\cite{yolo11}            & 86.9  & 25.3 & 52.3& 69.2& 55.7 & 6.23\\
YOLOv12-L~\cite{yolov12}           & 88.9  & 26.4 & 53.0  & 70.0  & 57.9 & 7.10\\ \rowcolor{blue!10}
\textbf{YOLOv13-L}& \textbf{88.4}  & \textbf{27.6} & \textbf{53.4}  & \textbf{70.9}  & \textbf{58.1} & \textbf{8.63}\\ \midrule
YOLOv8-X~\cite{yolov8}            & 257.8 & 68.2 & 54.0  & 71.0  & 58.8 & 12.83\\
RT-DETR-R101~\cite{rt_detr}        & 259.0 & 76.0 & 54.3  & 72.7  & –    & 13.51\\
RT-DETRv2-R101~\cite{rt_detrv2}      & 259.0 & 76.0 & 54.3  & 72.8  & –    & 13.51\\
YOLOv10-X~\cite{yolov10}           & 160.4     & 29.5    & 54.4& 71.3& 59.3& 10.70\\
YOLO11-X~\cite{yolo11}            & 194.9& 56.9& 54.2& 71.0& 59.1& 11.35\\
YOLOv12-X~\cite{yolov12}           & 199.0     & 59.1    & 54.4    & 71.1& 59.3& 12.46\\ \rowcolor{blue!10}
\textbf{YOLOv13-X}& \textbf{199.2} & \textbf{64.0} & \textbf{54.8}  & \textbf{72.0}  & \textbf{59.8} & \textbf{14.67}\\
\bottomrule
\end{tabular}
\vspace{-0.4cm}
\end{table*}

%% file: sections/figs/fig6-box.tex
\begin{figure*}
    \centering
    \includegraphics[width=1\linewidth]{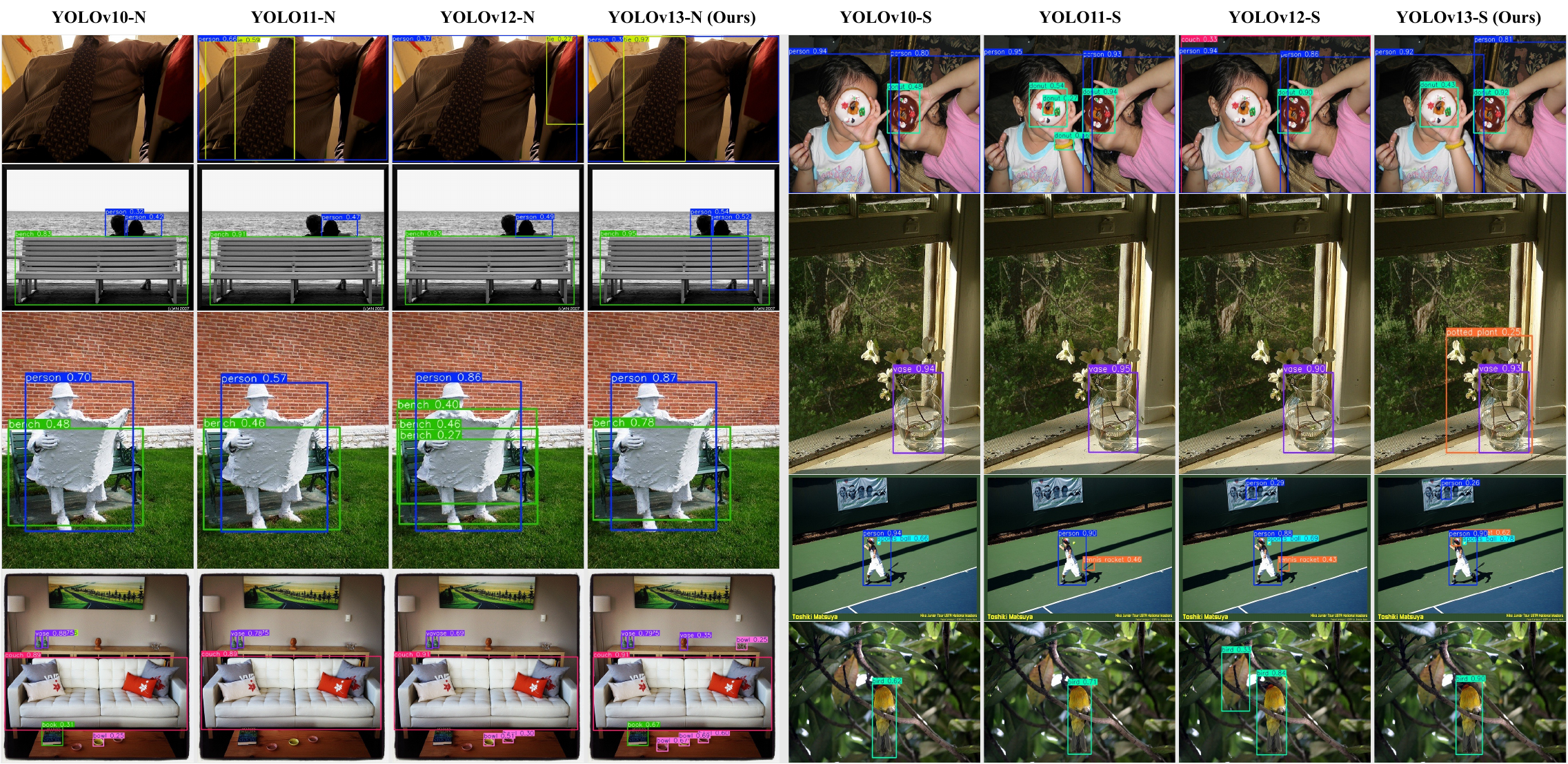}
    \vspace{-0.5cm}
    \caption{Qualitative comparison on MS COCO dataset~\cite{mscoco}. From left to right: detection results generated from YOLOv10-N/S~\cite{yolov10}, YOLO11-N/S~\cite{yolo11}, YOLOv12-N/S~\cite{yolov12}, and our YOLOv13-N/S models, respectively. Our method can achieve better detection performance in complex scenes. 
    }
    \vspace{-0.5cm}
    \label{fig:vis_bbox}
\end{figure*}

%% file: sections/tables/voc_comp.tex
\begin{table}[!tp]
\centering
\caption{Generalizability comparison of our method with other real-time object detectors. All models are trained on MS COCO dataset and tested on Pascal VOC 2007 dataset.}
\label{tab:voc_compare}
\renewcommand{\arraystretch}{1.15}
\setlength{\tabcolsep}{5.6mm}
\begin{tabular}{lccc}
\toprule
\textbf{Method} & \textbf{$\text{AP}_{50:95}^\text{val}$} & \textbf{$\text{AP}_{50}^\text{val}$} & \textbf{$\text{AP}_{75}^\text{val}$} \\
\midrule
YOLOv8-N~\cite{yolov8}            & 60.4& 80.4& 67.1\\
YOLOv10-N~\cite{yolov10}           & 61.8& 81.0& 68.3\\
YOLO11-N~\cite{yolo11}            & 62.1& 81.5& 68.9\\
YOLOv12-N~\cite{yolov12}           & 64.3& 83.0& 71.2\\
\rowcolor{blue!10}
\textbf{YOLOv13-N}        & \textbf{65.3}& \textbf{83.3}& \textbf{72.4}\\
\midrule
YOLOv8-S~\cite{yolov8}            & 66.4& 84.6& 72.4\\
YOLOv9-S~\cite{yolov9}            & 68.9& 85.8& 74.0\\
YOLOv10-S~\cite{yolov10}           & 67.7& 85.0& 75.9\\
YOLO11-S~\cite{yolo11}            & 67.7& 85.2& 74.8\\
YOLOv12-S~\cite{yolov12}           & 69.7& 86.5& 77.2\\
\rowcolor{blue!10}
\textbf{YOLOv13-S}       & \textbf{70.1} & \textbf{86.7}& \textbf{77.4}\\
\bottomrule
\end{tabular}
\vspace{-0.5cm}
\end{table}

%% file: sections/tables/distribution.tex
\begin{table}[!tp]
\centering
\setlength{\tabcolsep}{2.8mm}
\caption{Impact of setting different FullPAD distribution locations on the performance of our YOLOv13-S model.}
\vspace{-0.1cm}
\label{tab:distribution}
\renewcommand{\arraystretch}{1.0}
\begin{tabular}{ccccc}
\toprule
\textbf{Backbone-Neck} & \textbf{In-Neck} & \textbf{Neck-Head} & \textbf{$\text{AP}_{50:95}^\text{val}$} & \textbf{$\text{AP}_{50}^\text{val}$} \\
\midrule
\ding{56}  & \ding{56} & \ding{56} & 47.1& 64.1\\
\ding{52}  & \ding{56} & \ding{56} & 47.8& 65.0\\
\ding{56}  & \ding{52} & \ding{56} & 47.6& 64.8\\
\ding{56} & \ding{56} & \ding{52} & 47.7& 64.9\\
\ding{52} & \ding{52} & \ding{52} & 48.0& 65.2\\
\bottomrule
\end{tabular}
\vspace{-0.4cm}
\end{table}

%% file: sections/tables/num_edge.tex
\begin{table}[!tp]
\centering
\setlength{\tabcolsep}{2.6mm}
\caption{Impact of different numbers of hyperedges ($M$) on the performance of our YOLOv13-S model.}
\vspace{-0.1cm}
\label{tab:num_edge}
\renewcommand{\arraystretch}{1.0}
\begin{tabular}{cccccc}
\toprule
$M$ & \textbf{FLOPs (G)} & \textbf{Param. (M)} & \textbf{$\text{AP}_{50:95}^\text{val}$} & \textbf{$\text{AP}_{50}^\text{val}$} & \textbf{$\text{AP}_{75}^\text{val}$} \\
\midrule

2  & 20.4 & 8.6 & 47.8 & 64.7 & 51.9 \\
4  & 20.5& 8.8& 47.9& 65.0& 51.9\\
8  & 20.8 & 9.0 & 48.0 & 65.2 & 52.0 \\
16 & 21.5  & 9.6 &  48.1  & 65.2     &  52.1      \\
\bottomrule
\end{tabular}
\vspace{-0.4cm}
\end{table}

%% file: sections/tables/dsblocks.tex
\begin{table}[!tp]
\centering
\setlength{\tabcolsep}{1.2mm}
\caption{Impact of the DS blocks on the performance of our YOLOv13 Nano and Small models.}
\vspace{-0.1cm}
\label{tab:dsblocks}
\renewcommand{\arraystretch}{1.0}
\begin{tabular}{lccccc}
\toprule
\textbf{Method} & \textbf{DS Blocks} & \textbf{FLOPs (G)} & \textbf{Param. (M)} & \textbf{$\text{AP}_{50:95}^\text{val}$} & \textbf{$\text{AP}_{50}^\text{val}$} \\ \midrule
YOLOv13-N& \ding{52}  & 6.4  & 2.5  & 41.6 & 57.8 \\
YOLOv13-N& \ding{56}  & 7.9  & 3.1  & 41.6 & 57.9 \\ \midrule
YOLOv13-S& \ding{52}  & 20.8 & 9.0  & 48.0 & 65.2 \\
YOLOv13-S& \ding{56}  & 27.1 & 11.7 & 48.0 & 65.3 \\
\bottomrule
\end{tabular}
\vspace{-0.3cm}
\end{table}

%% file: sections/figs/fig7-edge.tex
\begin{figure}
    \centering
    \includegraphics[width=1\linewidth]{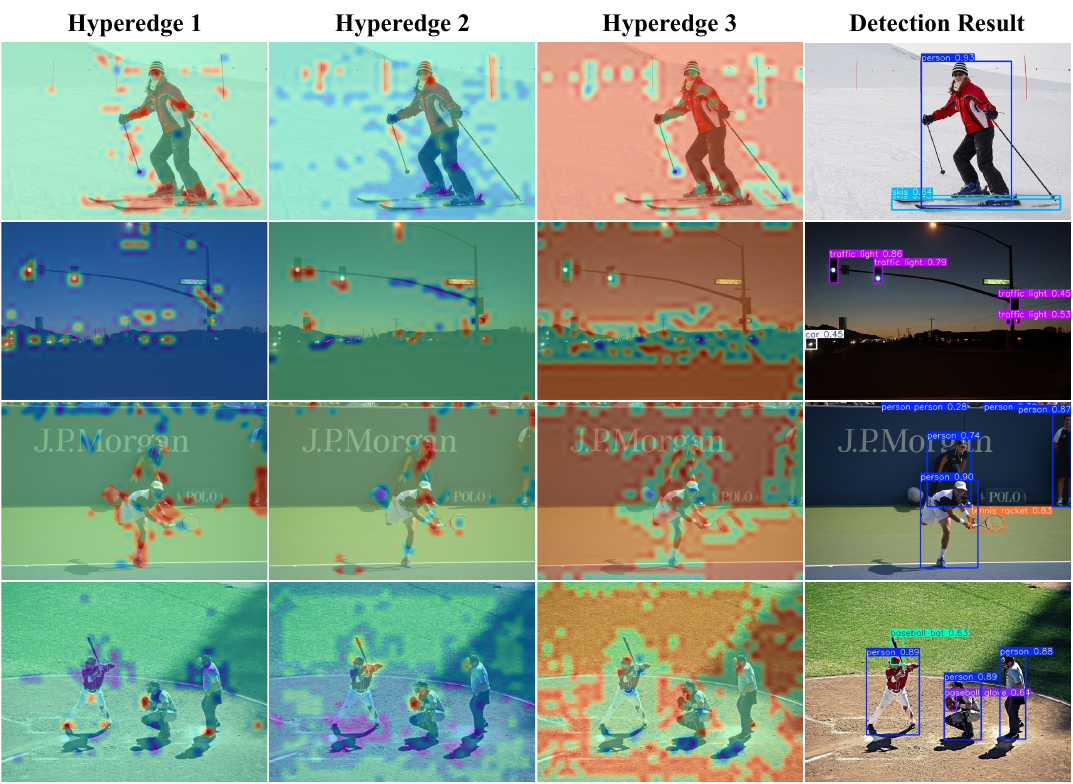}
    \vspace{-0.6cm}
    \caption{Visualization of adaptive hyperedges. The hyperedges in the first and second columns mainly focus on the high-order interactions among objects in the foreground. The third column mainly focuses on the interactions between the background and the foreground targets. The visualization of these adaptive hyperedges can intuitively reflect the high-order visual correlations modeled by our YOLOv13.}
    \vspace{-0.5cm}
    \label{fig:vis_soft}
\end{figure}

%% file: sections/tables/epoch.tex
\begin{table}[!tp]
\centering
\setlength{\tabcolsep}{2.6mm}
\caption{Impact of different training epochs on the performance of our YOLOv13 Nano and Small models.}
\label{tab:epoch}
\renewcommand{\arraystretch}{1.0}
\begin{tabular}{lcccc}
\toprule
\textbf{Epochs} & \textbf{$\text{AP}_{50:95}^\text{val}$ (N)} & \textbf{$\text{AP}_{50}^\text{val}$ (N)} & \textbf{$\text{AP}_{50:95}^\text{val}$ (S)} & \textbf{$\text{AP}_{50}^\text{val}$ (S)} \\
\midrule
300  & 39.5& 55.2& 46.9& 63.1\\
500  & 41.4& 57.6& 47.8& 64.9\\
600  & 41.6& 57.8& 48.0& 65.2\\
800 & 41.3& 57.3& 47.8& 64.7\\
\bottomrule
\end{tabular}
\vspace{-0.4cm}
\end{table}

%% file: sections/tables/latency.tex
\begin{table}[!tp]
\centering

\setlength{\tabcolsep}{3.8mm}
\caption{Inference latency (ms) for FP16 precision on various computing devices. }
\vspace{-0.2cm}
\label{tab:latency}
\renewcommand{\arraystretch}{1.0}
\begin{tabular}{lccc}
\toprule
\textbf{Method}           & \textbf{Tesla T4} & \textbf{RTX 4090} & \textbf{CPU (ONNX)}\\
\midrule
YOLOv13-N     & 1.97& 1.25& 39.97\\
YOLOv13-S    & 2.98& 1.27& 61.12\\
YOLOv13-L     & 8.63& 2.36& 185.36\\
YOLOv13-X    & 14.67& 3.10& 330.12\\
\bottomrule
\end{tabular}
\vspace{-0.3cm}
\end{table}


%% file: sections/6_conclusion.tex
\section{Conclusion}
In this paper, we propose YOLOv13, the state-of-the-art end-to-end real-time object detector. A hypergraph-based adaptive correlation enhancement mechanism is proposed to adaptively explore latent global high-order correlations and achieve multi-scale feature fusion and enhancement based on correlation guidance. The correlation-enhanced features are distributed throughout the entire network by the proposed full-pipeline aggregation-and-distribution paradigm, effectively promoting information flow and achieving accurate object detection. In addition, a series of blocks based on depthwise separable convolution are proposed, which can significantly reduce the number of parameters and FLOPs while maintaining accuracy. We conduct extensive experiments on the widely used MS COCO dataset. The quantitative and qualitative results demonstrate that our proposed method achieves state-of-the-art performance with lower computational complexity.